\RequirePackage{fix-cm}
\documentclass[twocolumn]{svjour3}          
\smartqed  
\usepackage{graphicx}
\usepackage{booktabs}
\usepackage{multirow}
\usepackage{bm}
\usepackage{subfigure}
\usepackage{booktabs}
\usepackage{amsfonts}
\usepackage{subfigure}
\usepackage{amsmath}
\usepackage{booktabs}
\usepackage{threeparttable}
\usepackage{latexsym}
\usepackage{algorithm}
\usepackage{algorithmic}
\usepackage[misc]{ifsym}
\usepackage[colorlinks,
linkcolor=red,
anchorcolor=blue,
citecolor=blue
]{hyperref}

\begin{document}
	
	\title{NCST: Neural-based Color Style Transfer for Video Retouching 
	}
	
	
	\author{Xintao Jiang$^{1,2}$        \and
		Yaosen Chen$^{2,3}$ \and
		Siqin Zhang$^{2}$ \and
		Wei Wang$^{2}$ \and
		Xuming Wen$^{2}$
	}


	\institute{
		{$^\textsubscript{\Letter}$}
		\\Xintao Jiang \\jiangxintao199@gmail.com
		\\Yaosen Chen\\chenyaosen@sobey.com
		\\
		\\
		1. Sichuan University-Pittsburgh Institute,Chengdu, Sichuan, 610065 China
		\\
		2. Sobey Media Intelligence Laboratory, Chengdu, Sichuan, 610041 China
		\\
		3. University of Electronic Science and Technology of China, Chengdu, Sichuan, 611731 China
	}

	\date{Received: date / Accepted: date}

	\maketitle
	
	\begin{abstract}
		Video color style transfer aims to transform the color style of an original video by using a reference style image. Most existing methods employ neural networks, which come with challenges like opaque transfer processes and limited user control over the outcomes. Typically, users cannot fine-tune the resulting images or videos. To tackle this issue, we introduce a method that predicts specific parameters for color style transfer using two images. Initially, we train a neural network to learn the corresponding color adjustment parameters. When applying style transfer to a video, we fine-tune the network with key frames from the video and the chosen style image, generating precise transformation parameters. These are then applied to convert the color style of both images and videos. Our experimental results demonstrate that our algorithm surpasses current methods in color style transfer quality. Moreover, each parameter in our method has a specific, interpretable meaning, enabling users to understand the color style transfer process and allowing them to perform manual fine-tuning if desired.
	\end{abstract}
	\begin{figure}[h] \centering \includegraphics[width=1\linewidth]{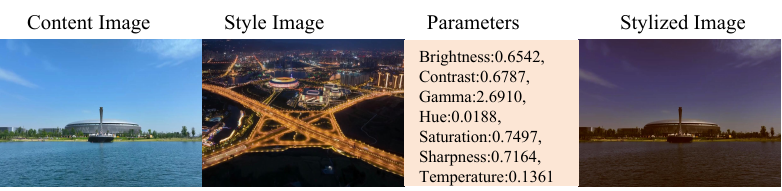} \caption{\textbf{Color Style Transfer.}  Given a content image and a style image, our method is able to transfer the color style from the style image to the content image with the color parameters.} \label{fig:tear} \end{figure}
	
	\section{Introduction}
	Image color style transfer involves transferring the color style of style image onto content image, thereby altering the color of the content image while preserving its original structure as shown in Figure~\ref{fig:tear}. Both image and video color style transfer have numerous real-world applications, including promotional material creation, post-production color grading of photos, and applying filters in videos and games to enhance thematic coherence and expressiveness. Consequently, this technology has attracted significant attention in recent years ~\cite{11,15}.
	
	Color style transfer, traditionally applied to individual images, is increasingly falling short of addressing the demands of video style transfer~\cite{10}. Techniques that rely on neural networks for image color style transfer may lead to disrupted visual continuity in videos. Moreover, these methods often result in inefficiencies due to inference with neural networks during the transfer~\cite{19}. 
	
	Another limitation of neural network-reliant color style transfer is its opacity—the inability of users to access or adjust color parameters during conversion, which makes post-adjustment fine-tuning impossible. The NLUT~\cite{16} method improves the efficiencies of color style transfer by predicting a 3D lookup table with neural networks. However, it still conceals the color adjustment parameters involved in the process. To address these challenges, we propose using neural networks to predict the parameters for color style transfer. Our method predicts color transfer parameters—such as contrast and brightness—between a style image and a content image, meeting thereby diverse color style transfer demands. Additionally, these parameters can be manually adjusted by users, offering greater flexibility. The parameters we predict are inherently suitable for video color adjustment, and when applied, they mitigate the flickering issues typical of neural network-based video color style transfer. Furthermore, our method supports test-on-time training for specific video transfers, enhancing its effectiveness for tailored applications. 
	
	These predicted color adjustment parameters can also be transformed into 3D LUTs for video color style transfer, achieving ultra-fast conversion speeds while maintaining high-quality results, similar to the capabilities of NLUT. In summary, our key contributions are as follows:
	
	\begin{itemize} 
		\item We propose a method that automatically predicts color style transfer parameters using a neural network for video style transfer, making the style transfer process transparent.
		\item Our method can be seamlessly integrated with other methods to achieve personalized objectives such as enhancing efficiency. 
		\item Experiments show that our method has superior performance in color style transfer effects and achieves higher consistency in video color style transfer. 
	\end{itemize}

	\section{Related Work}
	\textbf{Color Style Transfer.} As a critical area in computer vision, color style transfer has garnered widespread attention. This technology allows the transfer of color and style from one image to another while preserving the content and spatial structure of the original image.
	
	Early methods, such as Reinhard’s Color Transfer~\cite{10}, achieved color style transfer by adjusting the target image’s color distribution. However, this method often disregards spatial and structural information, leading to suboptimal results when transferring complex images or videos. To address these limitations, Gatys et al. \cite{11} proposed using convolutional neural networks for image style transfer, where high-level semantic features and low-level texture features are extracted via deep neural networks and applied to the target image. S. P. Premnath et al. \cite{27}proposed a method about Image enhancement and blur pixel identification. It can fix the blur and noise of stylized image. Du et al.\cite{26}proposed a new model which can greatly increase efficiency of transfer time, avoid distorted image structure and retain more detail information.These methods achieved notable success in image color style transfer but encountered temporal consistency issues when applied to videos, resulting in flickering and inconsistent styles across frames.
	
	Huang and Belongie \cite{12} addressed this problem by proposing Adaptive Instance Normalization (AdaIN), which enables fast and flexible style transfer by dynamically adjusting the statistics of convolutional feature maps. Johnson et al. \cite{13} further improved color style transfer quality through perceptual loss functions. In video color style transfer, Chen et al. \cite{14} alleviated temporal inconsistency by incorporating an optical flow-guided mechanism. Ruder et al. \cite{15} introduced temporal smoothing losses to neural networks, further enhancing inter-frame consistency. Wen and Li \cite{28}proposed a method called multi-scale LLVE can constraint inter-frame consistency in video without ground-truth labels.
	\begin{figure*}[h] \centering \includegraphics[width=1.0\linewidth]{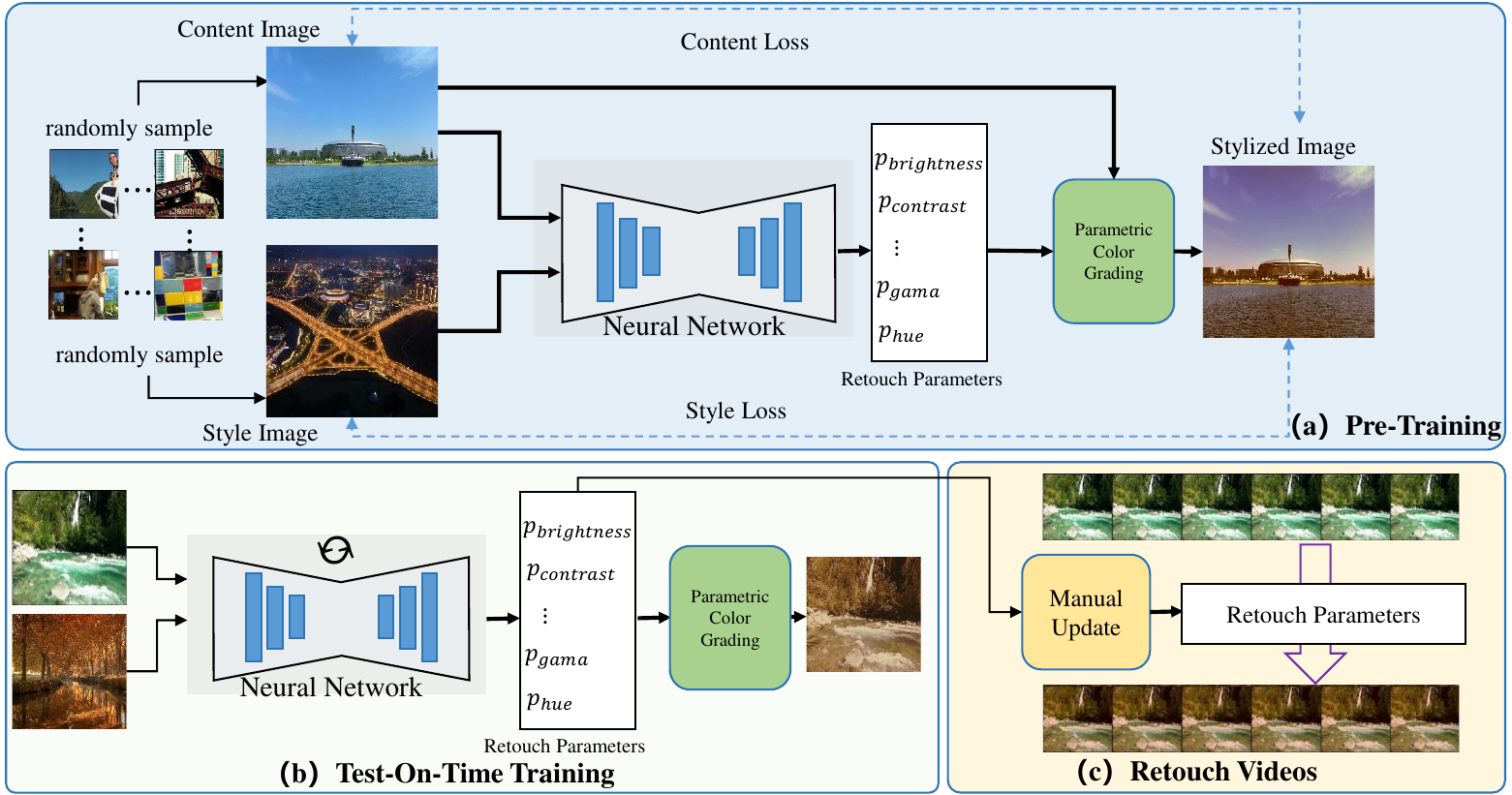} \caption{\textbf{An overview of neural-based color style transfer.} Our method consists of three steps: (a). Pre-Training, images are randomly selected from a large image dataset as style images and content images for pretraining to obtain a pretrained model. (b). Test-On-Time Training, the user's specific content video and style image are input into this pretrained model for fine-tuning, resulting in a specific color style transfer model used to generate parameters like brightness and contrast, forming a parameter set. (c). Retorch Videos, this parameter set is used to perform color style transfer on the content video, resulting in a stylized video.}\label{fig:overiew}
	\end{figure*}
	
	In addition to neural networks, 3D Lookup Tables (LUT) have been widely used for color style transfer. Chen et al. \cite{16} proposed a method for rapid color mapping using pre-generated LUTs, significantly improving video color style transfer efficiency. Despite improvements, challenges such as maintaining inter-frame consistency remain unresolved. Additionally, in previous methods, the entire color style transfer process was machine-determined, limiting the ability of users to make adjustments. To address these issues, we propose a neural-based color style transfer method that ensures both effectiveness and efficiency while enabling post-processing adjustments.
	
	\textbf{Parametric Color Grading.} Parametric color grading is a technique for altering the visual style of an image by modifying parameters like brightness, contrast, and hue. This method is extensively utilized in video post-production, photography, and digital art. Initially, users manually adjusted brightness and contrast, a process that demanded aesthetic judgment and expertise, making it inefficient for handling large volumes of images~\cite{23}. As a result, automatic color grading methods have gained popularity among professionals for their efficiency and ease of use.

	Earlier techniques allowed users to manipulate image parameters through color space transformations and curve adjustments \cite{10,24}. Boyadzhiev et al. \cite{25} introduced a user-guided local color editing method that enabled complex color grading effects with simple operations.
	
	Deep learning has since become widely adopted for color grading. Gatys et al. \cite{11} proposed a neural-based method that naturally fuses content and style. Johnson et al. \cite{22} achieved real-time style transfer through perceptual loss. He et al. \cite{20} used 3D Lookup Tables (LUTs) for color grading, simplifying complex color transformations into a set of adjustable parameters. While this method offers speed and ease of use, it may result in image blurring in some scenarios, and the predefined 3D LUTs may not adapt well to diverse environments and styles.
	
	Parametric color grading offers greater precision and control compared to LUTs. It typically involves multiple parameters, allowing users to make subtle adjustments to achieve the desired effect. Additionally, this method enables real-time color grading, significantly improving workflow efficiency by allowing users to preview adjustments immediately.
	
	However, parametric color grading also has its drawbacks. It can be more complex than LUT-based methods, making it challenging for users to start from scratch and achieve their desired effect. To address this, we propose a neural-based color style transfer method that generates color style transfer parameters automatically, helping users optimize parameter settings for their specific needs.
	\vspace{-5mm}

	\section{Our Method}
	\vspace{-3mm}
	An overview of neural-based color style transfer for video retouching is shown in Figure~\ref{fig:overiew}
	. We first randomly select two images from a large image dataset as the content image and style image, respectively. After inputting the content and style images into the neural network, we predict the color style transfer parameters. Using these parameters, we perform color style transfer on the content image to obtain the result image. We compare the result image with the content and style images to compute the color histogram matching loss, content loss, and style loss. The weighted sum of these losses is used as the total loss to pretrain the neural network model. By repeating this process, we obtain a pretrained model. The pretrained neural network model is then applied to the user's specific content and style images for further training and test-on-time training to predict specific color style transfer parameters. Finally, using these parameters, we perform color style transfer on the original video to obtain the target video. After obtaining the video, users can further adjust it based on the parameters provided by the neural network to achieve better results.
	
	\vspace{-5mm}
	\subsection{Color Grading Parameters Generation}
	
	We design a neural network into which we input the video content and the style image; the neural network outputs color style transfer parameters, which are used to perform color style transfer on the target video.
	
	First, we downscale the content and the style image that are to be input into the neural network to reduce subsequent memory usage and improve efficiency. Then, we extract the content and style features using a pretrained VGG network~\cite{17}, obtaining $F^{c}_{j}$ and $F^{s}_{j}$.

	Additionally, we perform convolutional feature extraction on the style and content features at different scales and fuse them using AdaIN~\cite{12}. Finally, we process the features fused by AdaIN using Adaptive Average Pooling (AvgPool)~\cite{18} to obtain features of the same dimension. We then concatenate these features to serve as parameters for color grading (also known as ``color style transfer"). In our method, each parameter corresponds to a different color adjustment technique, such as contrast or brightness. These techniques are combined to perform color style transfer on the video.
	
	\begin{figure*}[p] \centering \includegraphics[width=1\linewidth]{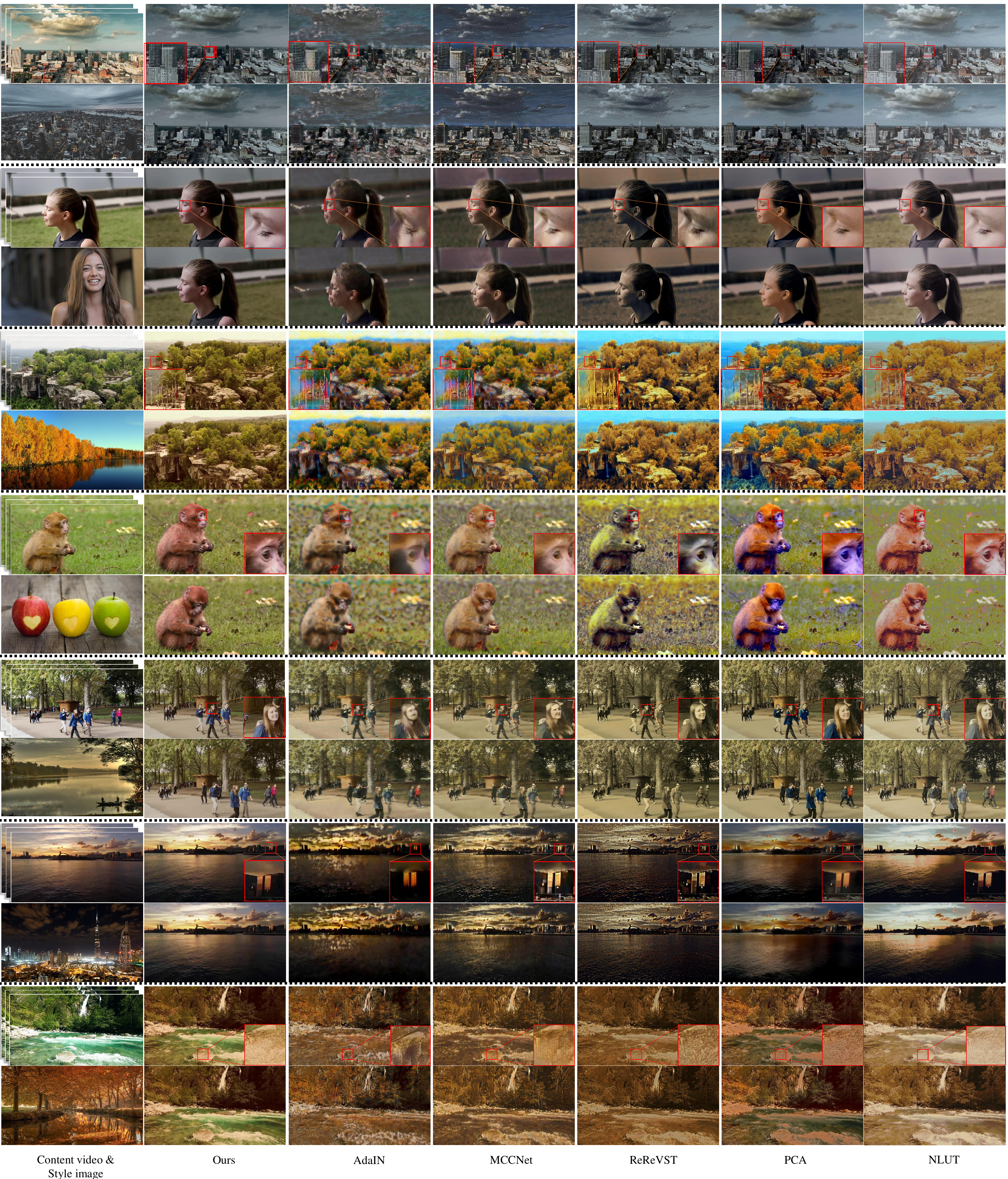} \caption{\textbf{Visual comparison of our method with other methods on multiple images.} These images can show our methods has better result compared with other methods } \label{fig:VisualComparison} 
	\end{figure*}

	\subsection{Loss Functions}
	
	We divide the loss into three parts: content loss, style loss, and color histogram matching loss. We use a pretrained VGG network~\cite{17} to extract features of content and style. For the style loss, we define it as:
	
	\begin{equation} L_s = \sum_{j=1}^{4} \left| \mu(F_j^s) - \mu(F_j^j) \right|_2 + \left| \sigma(F_j^s) - \sigma(F_j^j) \right|_2 \end{equation}
	where $F_j$ represents the feature map of the $j$-th layer extracted by the pretrained VGG model, and $\mu()$ and $\sigma()$ represent the mean and variance computed from the VGG network layers, respectively. $F^s$ denotes the style image, and $F^c$ denotes the stylized image. For the content loss, we use only the last layer feature map extracted by the pretrained VGG model:
	
	\begin{equation} L_c = \left| F_4^s - F_4^c \right|_2 \end{equation}
	
	For the color histogram loss, we define it as
	
	\begin{equation} L_{\text{color}} = \sum_{i=1}^{3} \sum_{j=1}^{N_{\text{bins}}} \hat{h}_{\text{target}, i, j} \cdot \log \left(\frac{\hat{h}_{\text{target}, i, j}}{\hat{h}_{\text{input}, i, j} + \epsilon} \right) \end{equation}
	where:
	\begin{equation} \hat{h}_{\text{input}, i, j} = \frac{h_{\text{input}, i, j} + \epsilon}{\sum_{k=1}^{N_{\text{bins}}} h_{\text{input}, i, k} + \epsilon} \end{equation}
	
	\begin{equation} \hat{h}_{\text{target}, i, j} = \frac{h_{\text{target}, i, j} + \epsilon}{\sum_{k=1}^{N_{\text{bins}}} h_{\text{target}, i, k} + \epsilon} \end{equation}
	where $h_{\text{input}, i, j}$ and $h_{\text{target}, i, j}$ represent the values in the $j$-th bin of the histogram for the $i$-th channel of the input image and target image, respectively. $N_{\text{bins}}$ is the number of bins in the histogram. $\epsilon$ is a very small positive number used to prevent division by zero and issues when computing the logarithm. 
	
	Finally, we combine the content loss, style loss, and color histogram loss with certain weights to obtain the total loss:
	
	\begin{equation} L = \lambda_s L_s + \lambda_c L_c + \lambda_{\text{color}} L_{\text{color}} \end{equation}
	where $\lambda_s$ is the weight of the style loss, $\lambda_c$ is the weight of the content loss, and $\lambda_{\text{color}}$ is the weight of the color histogram loss.

	\subsection{Training Strategy and Usage}
	
	We first pretrain the model on the large image dataset MS-COCO~\cite{19}. We randomly select images from the dataset as content images and style images, reducing the model's loss during the training process to achieve effective training. Next, we input the specific video and style image into the model. The model selects representative frames of the content video based on the color histogram and then fine-tunes the model using these representative frames and the style image to generate specific color style transfer parameters used for the color style transfer of the entire video.

	To apply this method for color style transfer in a specific video, we begin by inputting the user's content video and style image into the pretrained model for test-on-time training (as illustrated in Figure~\ref{fig:overiew} (b), Test-On-Time Training). This process generates a customized color style transfer model, which produces color grading parameters like brightness and contrast. These color grading parameters are then used to transform the color style of the content video, resulting in a stylized video (as shown in Figure~\ref{fig:overiew} (c), Retouch Videos).
	
	We assume that the predicted parameter is \textbf{P}, and the process of parametric color grading can be expressed as:
	\begin{equation} 
		[r^{'},g^{'},b^{b}] = f_{\textbf{P}}([r,g,b])
	\end{equation}
	where$[r,g,b]$, $[r^{'},g^{'},b^{'}]$represent the values of the pixels before and after color grading. The algorithm~\ref{algmultiple} shows color grading parameters and 3D LUT conversion. The color style transfer 3D LUT is obtained by equispaced sampling in 8-bit RGB space, and each sampled color value is calculated and normalized by using prediction parameters. Where $N$ is the number of samples and $f$ is the color transfer calculation of the color values using the predicted color style transfer parameters.
	\begin{algorithm}
		\caption{Color grading parameters and 3D LUT conversion}
		\begin{algorithmic}
			\FOR {$b\in [0,N]$}
			\FOR {$g\in [0,N]$}
			\FOR {$r\in [0,N]$}
			\STATE $LUT[i]=\frac{f([r,g,b]\times \frac{255}{N})}{255}$
			\ENDFOR
			\ENDFOR
			\ENDFOR
		\end{algorithmic}
		\label{algmultiple}
		
	\end{algorithm}
	\vspace{-8mm}
	\section{Experiments}
	We conducted both qualitative and quantitative tests on our method and compared it with other advanced color style transfer methods. In the pretraining stage, we used the Adam optimizer~\cite{1} with a batch size of 6 to train the neural network for 20,000 iterations. In the test-on-time training stage, we used a batch size of 1 and performed 500 iterations, which took approximately 1 minute. The learning rates for both pretraining and test-on-time training were the same, set at $10^{-4}$. We used images from MS-COCO as the training set during the pretraining phase of our method. During pretraining, the content and style images were resized to $256 \times 256$ pixels. All experiments were conducted on a single NVIDIA GeForce RTX 3070 Ti Laptop GPU. The videos used for testing were sourced from Pixabay\footnote{https://pixabay.com/videos} and Pexels\footnote{https://www.pexels.com/}, each with specifications of 25 frames per second and a duration of 6 seconds.

	\begin{table}[h]
		\centering
		\resizebox{0.5\textwidth}{!}{
			\begin{tabular}{l|c|c|c|c|c|c|c}
				\hline
				\textbf{Method} & \textbf{city} & \textbf{girl} & \textbf{kelly} & \textbf{monkey} & \textbf{pedestrian} & \textbf{sunset} & \textbf{Average} \\
				\hline
				AdaIN & 0.0047 & 0.0029 & 0.0096 & 0.0028 & 0.0057 & 0.0025 & 0.0052 \\
				MCCNet & 0.0075 & 0.0046 & 0.0055 & 0.0020 & 0.0045 & 0.0015 & 0.0034 \\
				ReReVST & 0.0028 & 0.0032 & 0.0078 & 0.0032 & 0.0032 & 0.0027 & 0.0042 \\
				PCA & 0.0062 & 0.0039 & 0.0077 & 0.0102 & 0.0118 & 0.0017 & 0.0078 \\
				NLUT & \textbf{0.0030} & \textbf{0.0018} & \textbf{0.0044} & \textbf{0.0018} & \textbf{0.0035} & \textbf{0.0005} & \textbf{0.0026} \\
				Ours & \textbf{0.0033} & \textbf{0.0024} & \textbf{0.0053} & \textbf{0.0015} & \textbf{0.0035} & \textbf{0.0004} & \textbf{0.0027} \\
				\hline
		\end{tabular}}
		\caption{Long-term consistency. We use the warped error (↓) to compare long-term consistency. The best results are highlighted.}
		\vspace{-8mm}
		\label{tab:advanced_metrics}
	\end{table}
	\begin{table}[h]
		\centering
		\resizebox{0.5\textwidth}{!}{
			\begin{tabular}{l|c|c|c|c|c|c|c}
				\hline
				\textbf{Method} & \textbf{city} & \textbf{girl} & \textbf{kelly} & \textbf{monkey} & \textbf{pedestrian} & \textbf{sunset} & \textbf{Average} \\
				\hline
				AdaIN & 0.0019 & 0.0004 & 0.0044 & 0.0012 & 0.0023 & 0.0020 & 0.0025 \\
				MCCNet & 0.0033 & 0.0003 & 0.0024 & 0.0007 & 0.0019 & 0.0015 & 0.0016 \\
				ReReVST & 0.0010 & 0.0002 & 0.0015 & 0.0014 & \textbf{0.0013} & 0.0009 & 0.0013 \\
				PCA & 0.0011 & 0.0002 & 0.0010 & 0.0009 & 0.0021 & 0.0022 & 0.0015 \\
				NLUT & \textbf{0.0008} & \textbf{0.0001} & 0.0008 & 0.0006 & 0.0015 & 0.0015 & 0.0011 \\
				Ours & 0.0009 & \textbf{0.0001} & \textbf{0.0008} & \textbf{0.0005} & 0.0015 & \textbf{0.0007} & \textbf{0.0007} \\
				\hline
		\end{tabular}}
		\caption{
			Short-term consistency. We use the warped error (↓) to compare short-term consistency. The best results are highlighted.
		}
		\vspace{0mm}
		\label{tab:metrics}
	\end{table}
	
	\begin{table}[h]
		\centering
		\resizebox{0.5\textwidth}{!}{
			\begin{tabular}{l|c|c|c|c|c|c|c|c}
				\hline
				\textbf{Method} & \textbf{512} & \textbf{HD} & \textbf{FHD} & \textbf{QHD} & \textbf{2000} & \textbf{4K} & \textbf{5K} & \textbf{8K} \\
				\hline
				AdaIN & 36.28 & 118.32 & 255.19 & 412.26 & 444.68 & 1025.05 & 1832.61 & OOM \\
				MCCNet & 63.83 & 209.81 & 465.52 & 807.26 & 913.65 & 2045.06 & OOM & OOM \\
				ReReVST & 29.58 & 101.25 & 221.86 & 409.54 & 471.02 & 980.15 & OOM & OOM \\
				PCA & 65.76 & 74.50 & 101.99 & 186.81 & 198.48 & 381.22 & 669.99 & OOM \\
				NLUT & \textbf{0.05} & \textbf{0.05} & \textbf{0.10} & \textbf{0.19} & \textbf{0.27} & \textbf{0.43} & \textbf{0.76} & \textbf{1.72}\\
				Ours (LUT) & \textbf{0.34} & \textbf{0.76} & \textbf{1.31} & \textbf{2.53} & \textbf{2.76} & \textbf{6.65} & \textbf{9.59} & \textbf{21.25} \\
				Ours & 37.38 & 53.01 & 77.42 & 144.91 & 150.75 & 287.86 & 521.27 & OOM \\
				\hline
		\end{tabular}}
		\caption{Efficiency comparison: We use ``milliseconds/frame'' (↓) to compare efficiency at different resolutions. The best results are highlighted.}
		\vspace{-8mm}
		\label{tab:EfficiencyComparison}
	\end{table}
	\vspace{-8mm}
	\subsection{Qualitative Results}
	Our method achieves excellent results in image color style transfer. We compared our method with the following five methods: PCA~\cite{ref4}, AdaIN~\cite{12}, MCCNet~\cite{ref5}, ReReVST~\cite{ref26} and NLUT~\cite{16}.
	
	As shown in Figure~\ref{fig:VisualComparison}, overall, the images generated by our method are closer to real scenes. When zoomed in, it is evident that other methods have issues such as blurring or incorrect colors in details, whereas our method, by directly using parameters for color adjustment, almost does not have such problems. Moreover, our method allows for further adjustment of parameters after color style transfer to achieve better results. Due to the limited number of adjustable parameters in the method and the lack of finer color adjustment capabilities, it is challenging to achieve color style transfer effects consistent with traditional methods by directly adjusting the existing parameters.

	\subsection{Quantitative Results}
	
	\textbf{Consistency Measurement}. To address the flickering and consistency issues that occur after video color style transfer, we choose to directly measure numerical values to demonstrate the superiority of our method. We used the warped Learned Perceptual Image Patch Similarity (LPIPS) metric~\cite{35} to measure both long-term and short-term consistency of videos. In testing long-term consistency, we extracted one frame every 35 frames of the video for testing. In testing short-term consistency, we extracted one frame every 5 frames. The test results are shown in Table~\ref{tab:advanced_metrics}
	and Table~\ref{tab:metrics}.
	
	It can be seen that our method outperforms other methods in short-term consistency and is slightly weaker than NLUT in long-term consistency in some scenes, but the overall effect is comparable. Moreover, our method generates parameters that can be manually fine-tuned later to achieve better results.

	\begin{figure*}[htbp] \centering \includegraphics[width=1\linewidth]{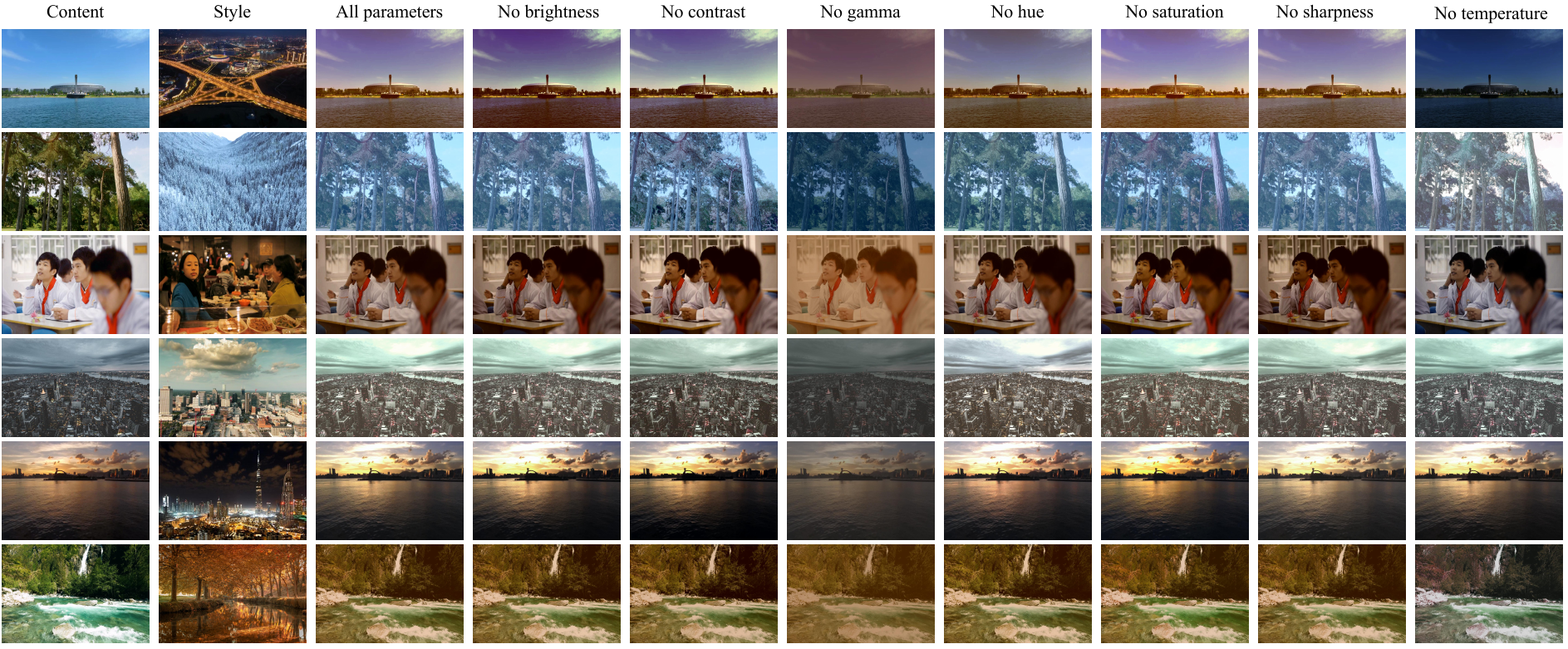} \caption{\textbf{The impact of various color grading parameters on the final color grading effect.} This experiment demonstrates the impact of removing individual adjustment parameters on style transfer results. The results show that each parameter contributes differently to the final effect.} \label{fig:ab_removing} \end{figure*}

	\begin{figure}[h] \centering \includegraphics[width=0.8\linewidth]{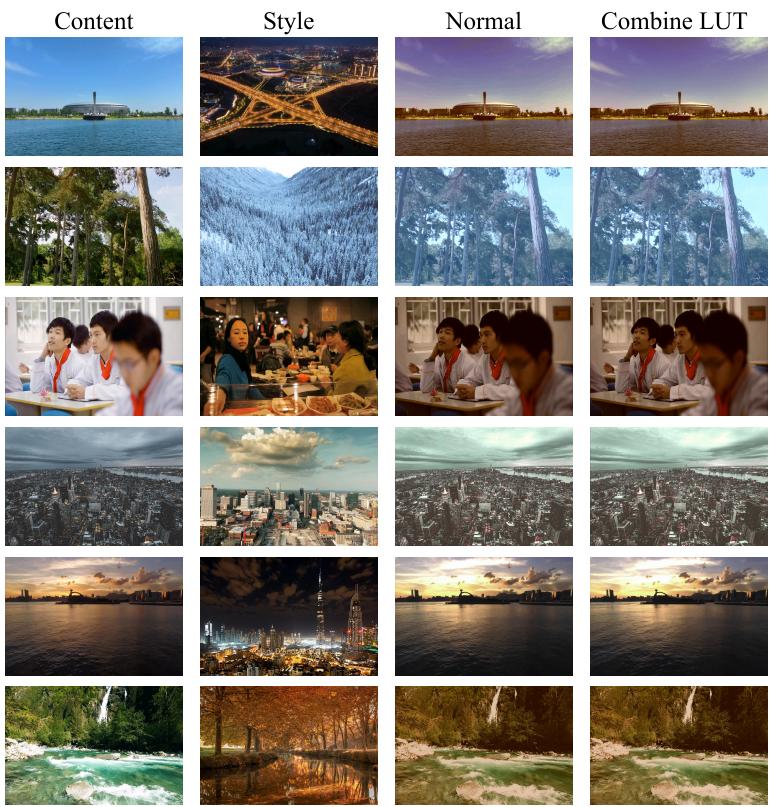} \caption{\textbf{The impact of converting color grading parameters into 3D LUT on color grading effects.} In this experiment, we predicted color grading parameters while generating a 3D LUT and applied it in DaVinci Resolve. The ``Content" refers to the content images, ``Style" refers to the style images, ``Normal" refers to results obtained directly using our method, and ``Combine LUT" shows the effects of applying the generated LUT in DaVinci Resolve.} \label{fig:CombineLUT} \end{figure}

	\begin{figure*}[h] \centering \includegraphics[width=1\linewidth]{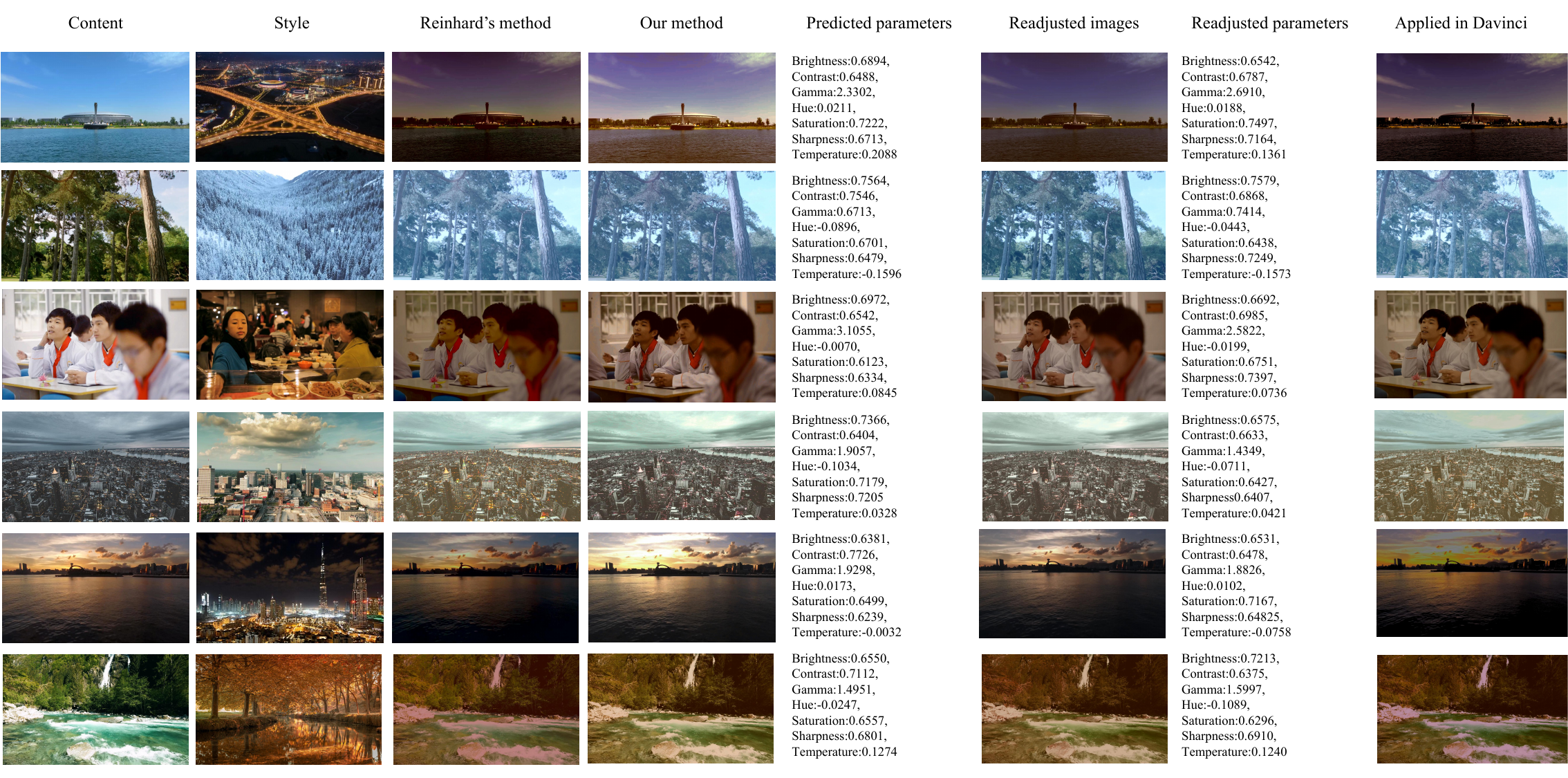} \caption{\textbf{The impact of manual color grading on the final result.} This figure compares the style transfer effects of a traditional method and our method. ``Reinhard's method" performs style transfer by calculating the color mean differences and standard deviation ratios between the content and style images. ``Our method" refers to the results obtained directly using our method. ``Predicted Parameters" are the color style transfer parameters derived from our method.``Readjusted Images" and ``Readjusted parameters" are the results and parameters after manual color grading. ``Applied in DaVinci" shows our method's results adjusted in DaVinci Resolve to resemble Reinhard's method.} \label{fig:Reinhard} \end{figure*}

	\begin{figure}[htbp] \centering \includegraphics[width=1\linewidth]{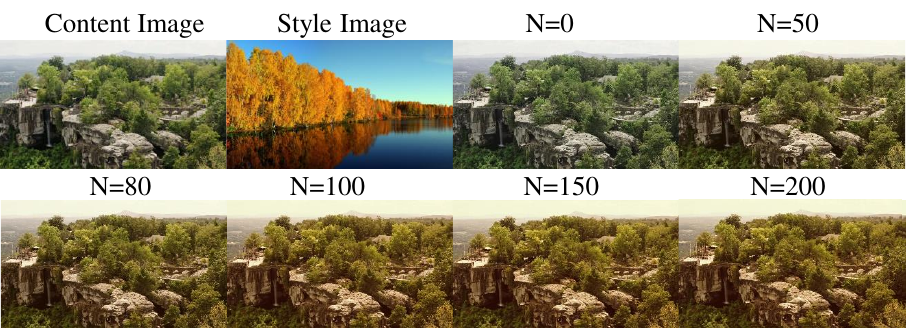} \caption{\textbf{The impact of Test-On-Time Training on video retouching.} } \label{fig:pretrain} \end{figure}

	\begin{figure}[htbp] \centering \includegraphics[width=1\linewidth]{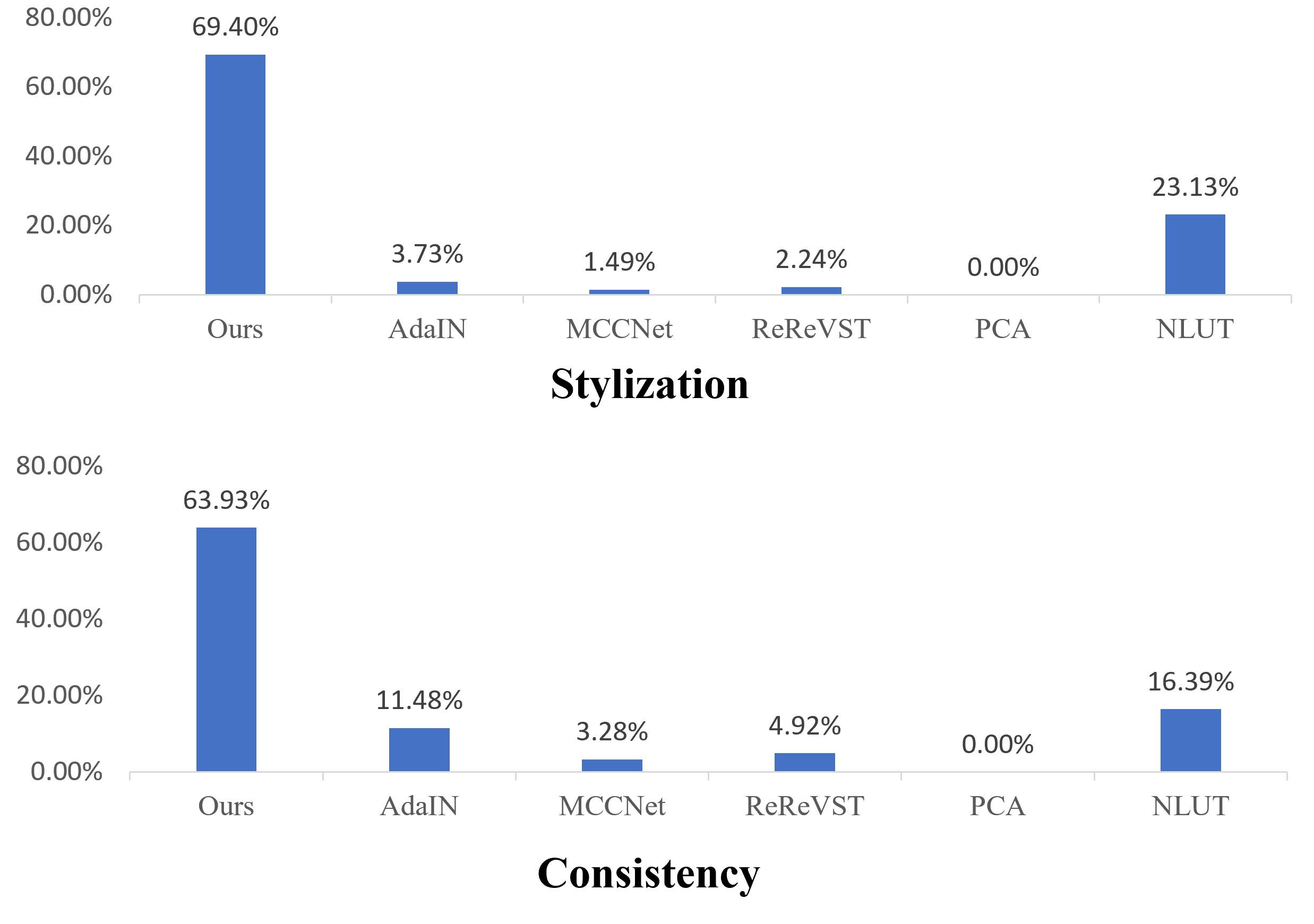} \caption{\textbf{Userstudy.}This bar graph shows that our method has higher recognition compared with other methods in user study. } \label{userstudy} \end{figure}

	\textbf{Efficiency Comparison.}	In Table~\ref{tab:EfficiencyComparison}, we compare the efficiency of video generation by calculating the time required to generate each frame. We compared generation efficiency at different resolutions: $512 \times 512$ (512), $1280 \times 720$ (HD), $1920 \times 1080$ (FHD), $2560 \times 1440$ (QHD), $2000 \times 2000$ (2000), $3840 \times 2160$ (4K), $5120 \times 2880$ (5K), and $7680 \times 4320$ (8K). We used image transformation code in PyTorch with \texttt{torchvision.transforms} to transform images indicated as ``Ours". We also converted the color grading parameters into a 3D LUT, indicated as ``Ours (LUT)", and tested it. The results show that our method outperforms other methods, except NLUT, for color style transfer. As video resolution increases, the increase in frame generation time for our method is smaller compared to other methods. The conversion of color grading parameters into a 3D LUT with our method effectively solves the memory usage problem in high-resolution image generation and significantly increases the efficiency of video generation. However, there still exists a gap in efficiency compared to NLUT, likely due to the lack of CUDA optimization when using LUT for image processing.
	
	\textbf{User study.} In our study, we conducted a user evaluation to compare our method with state-of-the-art methods focusing on stylization and consistency. We applied six different stylization methods, PCA~\cite{ref4}, AdaIN~\cite{12}, MCCNet~\cite{ref5}, ReReVST~\cite{ref26}, and NLUT~\cite{16}, to a set of six videos. Sixty participants were then invited to participate in the evaluation. Initially, we presented the participants with content and style images. Subsequently, they were shown the results of the various color style transfer methods. Participants were asked to select the method they deemed best in terms of stylization and consistency. Their preferences were recorded and analyzed, as illustrated in Figure~\ref{userstudy}. The bar graph clearly demonstrates that our method surpasses others in both stylization and consistency.

	\subsection{Ablation Study}
	
	\textbf{The impact of various color grading parameters on the final color grading effect.} In our experiments, we used the color grading parameters ``{brightness}, {contrast}, {gamma}, {hue}, {saturation}, {sharpness} and {temperature}". However, intuitively, not all color ``styles" require all of these parameters to express. Therefore, we investigate how each parameter influences the method's outcome to verify this point of view. We obtained the tested models by omitting a certain prediction parameter for test-on-time training. These models are used to perform style transfer and compare the generated images. The results are shown in Figure ~\ref{fig:ab_removing}. It is obvious that each parameter has a unique impact on color style transfer. The importance of different parameters varies in different images, which shows that users can control the style transfer effect by adjusting the parameters. This flexibility allows to achieve the desired results and obtain more diverse and dynamic style transfer effects.
	
	\textbf{The impact of converting color grading parameters into LUT on color grading effects.} 
	The color grading parameters converted to 3D LUT can be easily used in other commercial software, such as DaVinci Resolve\footnote{https://www.blackmagicdesign.com/products/davinciresolve.}. Therefore, we verified the effect comparison between direct color grading with the predicted color grading parameters and color grading with 3D LUT on DaVinci Resolve in Figure~\ref{fig:CombineLUT}. From the results, there is no difference between them, so our method can be easily integrated into other products.

	
	\textbf{The impact of manual color grading on the final result.} Due to limited adjustment parameters and the lack of finer color grading capabilities, it's challenging to achieve traditional style transfer effects by directly adjusting existing parameters. We combined the predicted color grading parameters with manual color grading to verify the feasibility of this combination of automatic and manual color grading. Figure~\ref{fig:Reinhard} shows that although the color grading effect obtained by directly obtaining the parameters is not as good as that of the Reinhard~\cite{10} method, our method can flexibly connect to manual modification of parameters to achieve better color grading, and this method is easier to connect to commercial software such as DaVinci Resolve.

	\textbf{The impact of Test-On-Time Training on video retouching.} Although the pre-trained model performs well in most content images, it is not effective for all content images. For example, the effect of the content image in Figure~\ref{fig:pretrain} is not very obvious. In this case, we can correct the performance of the model in this case through test-on-time training. We set to 0, 100, 200, 300, 400, and 500, respectively. It can be seen that when the number of pretraining iterations is between 0 and 300, the final generated results are not ideal. However, when the number of pretraining iterations reaches 300 or more, the converted images achieve relatively good effects. This indicates that pretraining the neural network is effective and can greatly increase the stability of the method. Experiments show that pretraining for 500 iterations takes approximately 1 minute, requiring only a small amount of time to significantly enhance the model's stability.

	

	\vspace{-5mm}

	\section{Conclusion}
	\vspace{-3mm}
	We propose a method that uses neural networks to generate parameters for color style transfer. This method combines neural network application with parametric color grading. Our method effectively addresses inter-frame consistency issues in video style transfer. When integrated with other techniques, it can enhance efficiency and fulfill specific style transfer requirements and other personalized requirements. However, our method has limitations; for videos with highly variable styles, it may be necessary to segment the video and perform multiple test-on-time training sessions to generate different color style transfer parameters, ensuring consistent results results. In the future, we will continue to explore combining our method with others to improve inter-frame consistency in color style transfer and achieve better outcomes.

	%
	
	\section*{Ethical and informed consent for data used}
	All data used came from public datasets. No additional personal data was collected.
	\section*{Data availability and access}
	The MS-COCO~\cite{19} dataset is available at: \\
	https://cocodataset.org/.

\end{document}